\documentclass[conference]{IEEEtran}

\IEEEoverridecommandlockouts                              

\usepackage{color}

\usepackage{times}
\usepackage{epsfig}
\usepackage{graphicx}
\usepackage{tabularx}
\usepackage{amsmath}
\usepackage{amssymb}
\usepackage{bbm}
\usepackage[misc]{ifsym}
\usepackage[export]{adjustbox}
\usepackage{enumitem}

\usepackage[ruled,vlined,linesnumbered]{algorithm2e}
\usepackage{graphicx} 

\usepackage{subfigure}
\usepackage{tikz}
\usepackage{verbatim}
\usepackage{booktabs}
\usepackage{multirow}
\usepackage{makecell}
\usepackage{authblk}
\usepackage{hyphenat} 

\usepackage{pgf}
\usepackage{algpseudocode}
\usetikzlibrary{arrows,automata}
\makeatletter
\let\NAT@parse\undefined
\makeatother
\usepackage[colorlinks]{hyperref}  

\title{\LARGE \bf
Hierarchical Needs Based Self-Adaptive Framework For Cooperative Multi-Robot System}

\author{Qin Yang \and Ramviyas Parasuraman\thanks{* The authors are with the Heterogeneous Robotics (HeRo) Lab, Department of Computer Science, University of Georgia, Athens, GA 30602, USA. 

Email: {\it \{qy03103,ramviyas\}@uga.edu}. }}

\begin{document}

\newtheorem{definition}{Definition}
\newtheorem{theorem}{Theorem}
\newtheorem{lemma}{Lemma}
\newtheorem{proposition}{Proposition}
\newtheorem{property}{Property}
\newtheorem{observation}{Observation}
\newtheorem{corollary}{Corollary}

\maketitle
\thispagestyle{empty}
\pagestyle{empty}

\begin{abstract}
    
Research in multi-robot and swarm systems has seen significant interest in cooperation of agents in complex and dynamic environments. 
To effectively adapt to unknown environments and maximize the utility of the group, robots need to cooperate, share information, and make a suitable plan according to the specific scenario. 
Inspired by Maslow's hierarchy of human needs and systems theory, we introduce {\it Robot's Need Hierarchy} and propose a new solution called {\it Self-Adaptive Swarm System} (SASS). It combines multi-robot perception, communication, planning, and execution with the cooperative management of conflicts through a distributed {\it Negotiation-Agreement Mechanism} that prioritizes robot's needs.
We also decompose the complex tasks into simple executable behaviors through several {\it Atomic Operations}, such as selection, formation, and routing.
We evaluate SASS through simulating static and dynamic tasks and comparing them with the state-of-the-art collision-aware task assignment method integrated into our framework.

\end{abstract}
\section{Introduction}
Natural systems (living beings) and artificial systems (robotic agents) are characterized by apparently complex behaviors that emerge as a result of often nonlinear spatiotemporal interactions among a large number of components  at different levels of organization \cite{levin1998ecosystems}.  Simple principles acting at the agent level can result in complex behavior at the global level in a swarm system. 
Swarm intelligence is the collective behavior of distributed and self-organized systems \cite{hamann2008framework}. 


Multi-robot systems (MRS) \cite{parker2008multiple} potentially share the properties of swarm intelligence in practical applications such as search, rescue, mining, map construction, exploration. MRS that allows task-dependent dynamic reconfiguration into a team is among the grand challenges in Robotics \cite{yang2018grand}, necessitating the research at the intersection of communication, control, and perception. 
Currently, planning-based approaches combined with star-shaped communication models can not generally scale or handle a large number of agents in a distributed or decentralized manner \cite{desaraju2012decentralized}. 

\begin{figure}[tbp]
\centering
\includegraphics[width=0.45\textwidth]{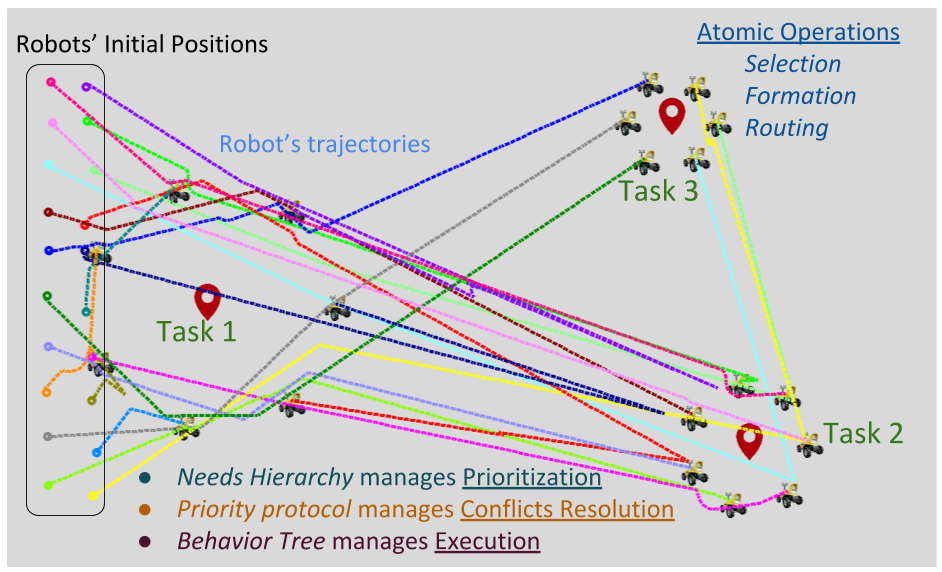}
\caption{Illustration of reactive multi-robot planning where robots move to Task 1 and Task 2 from their initial positions. During this execution, Task 3 is assigned and the robots react to this new task requirements.}
\label{fig: overview}
\end{figure}

Rizk \cite{rizk2019cooperative} group heterogeneous MRS into four levels based on the task complexity and the level of automation. The fourth level of automation combines task decomposition, coalition formation, task allocation, and task execution (planning) and control. However, researches addressing this fourth level is very thin.
Therefore, to drive advanced automation systems, the MRS framework needs to be combined with auxiliary controllers to handle conflicts, decompose the complicated task, and adapt to dynamic changes in the task assignments.

In this paper, we propose a MRS cooperation concept, Self-Adaptive Swarm System (SASS)\footnote{We use the term "Swarm" to denote the multi-agent context of the proposed multi-robot cooperation framework.}, that combines the parts of the perception, communication, planning, and execution to address this gap. See Fig.~\ref{fig: overview} for an illustration of the concept.
Our preliminary work published as an extended abstract in \cite{mrs2019} introduced the problem of multi-robots fulfilling dynamic tasks using state transitions represented through a Behavior Tree (BT) \cite{colledanchise2018behavior} and laid the foundations for the contributions made in this paper, which are outlined below.

\begin{figure}[t]
\centering
\includegraphics[width=0.3\textwidth]{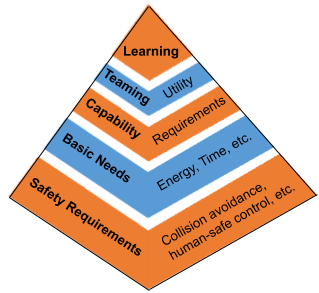}
\caption{Hierarchy of Robot Needs.}
\label{fig: needs}
\vspace{-4mm}
\end{figure}

\begin{itemize}[leftmargin=*]
    \item \textbf{Robot's Need Hierarchy:} To model an individual robot's $motivation$ and $needs$ in the negotiation process, we introduce the prioritization technique inspired by Maslow's hierarchy of human psychological needs \cite{maslow1943theory} and solve the conflicts associated with the sub-tasks/elements in task planning. We define the \textit{Robot's Need Hierarchy} at five different levels (see Fig. \ref{fig: needs}): {safety needs; basic needs (energy \cite{parasuraman2012energy}, time constraints, etc.); capability (heterogeneity, hardware differences, communication, etc.); team cooperation (global utility, team performance, cooperation, and global behaviors); and self-upgrade (learning)}.
    
    \item \textbf{Negotiation-Agreement Mechanism}: We propose a \textit{distributed  Negotiation-Agreement} mechanism for selection (task assignment), formation (shape control), and routing (path planning) in MRS, represented through a BT \cite{mrs2019} for automated planning of state-action sequences.
    
    \item \textbf{Atomic Operation:} We decompose the complex tasks into a series of simple sub-tasks through which we can recursively achieve those sub-tasks until we complete the high-level task. We provide several \textit{Atomic Operations} for the swarm behavior: $ Selection $, $ Formation $, and $ Routing $, which allows us to decompose a particular robot's action plans as flocking, pattern formation, and route planning under the same framework.
\end{itemize}

\section{Related Work}

Swarm robotics and swarm intelligence have been well studied in the literature \cite{hamann2008framework}. 
Multi-robot modeling and planning algorithms are among those well-studied topics yet require task-specific or scenario-specific application limitations. Martinoli \cite{martinoli1999swarm} presents the modeling technique based on rate equations, a promising method using temporal logic to specify and possibly prove emergent swarm behavior by Winfield et al. \cite{winfield2005formal}. Soysal and Sahin \cite{soysal2006macroscopic} apply combinatorics, and linear algebra is deriving a model for an aggregation behavior of swarms. Some studies also applied control theory to model and analyzed multi-robot and swarm systems \cite{gazi2003stability,feddema2002decentralized}. Recently, Otte et al. \cite{otte2019auctions} discussed various auction methods for multi-robot task allocation problem in communication-limited scenarios where the rate of message loss between the auctioneer and the bidders are uncertain.

From the multi-agent systems perspective, one of the earliest pioneering works, especially in the distributed artificial intelligence, includes \cite{smith1981frameworks}, where the authors defined the Contract Net Protocol (CNP) for decentralized task allocation. Aknine \cite{aknine2004extended} extended this idea to $m$ manager agents and $n$ contractor agents negotiation. A protocol for dynamic task assignment (DynCNET) has been developed by Weyns \cite{weyns2007dyncnet}. However, these methods rely on a central agent (such as an auctioneer or a contractor) to design the negotiation protocol and a supportive framework. They do not generally consider the changes in the agents' status, which restricts the direct applicability in real-world scenarios but only with adaptation. 

Importantly, unlike distributed robotic systems, MRS emphasizes a large number of agents and promotes scalability, for instance, by using local communication \cite{hamann2018swarm}, which plays a vital role in the whole system building various relationships between each robot to adapt to different environments and situations.
In Tab.~\ref{tab:Comparision}, we present a comparison of SASS against common MRS frameworks that focus on task allocation and planning at different automation levels.

In multi-robot planning and control, several studies focus on navigating a robot from an initial state to a goal state \cite{wagner2015subdimensional,ayanian2010abstractions,wu2019collision}. So the individual robot's entire route planning is usually computed in high-dimensional joint configuration space.
Since formations require robots to maintain stricter relative positions as they move through the environment, the flocking problem could be viewed as a sub-case of the formation control problem, requiring robots to move only minimal requirements for paths taken by specific robots \cite{olfati2006flocking}.
The research question here is how to design suitable local control laws for each robot to complete the globally assigned tasks efficiently and cooperatively and how paths can be planned for permutation-invariant multi-robot formations \cite{kloder2006path}.
Earlier, the solutions for such problems in flocking and formations are based on local interaction rules \cite{reynolds1987flocks} or behavior-based approaches \cite{mataric1993designing,balch1998behavior}. More recent approaches focus on proving stability and convergence properties in multi-robot behaviors basing on control-theoretic principles \cite{cao2012overview,murray2007recent,parasuraman2018multipoint}. 

To summarize, we aim to develop more advanced MRS by seeking the fourth level of the autonomous system, which combines task decomposition, group formation, planning, and control \cite{rizk2019cooperative}. More importantly, MRS should be able to adapt to the dynamic changes in the environment and task assignments.


\begin{table*}[ht]
\caption{Comparison of typical multi-robot system frameworks in the literature.}
\label{tab:Comparision}
\centering
\resizebox{\textwidth}{!}{\begin{tabular}{|c|c|c|c|c|c|c|c|c|c|c|c|c|c|c|}
\hline 
\multirow{2}*{Approach} & \multirow{2}*{Ref.} & \multirow{2}*{HET} & \multirow{2}*{COOP} & \multirow{2}*{COM} & \multirow{2}*{NEGO} & \multirow{2}*{DEC} & \multirow{2}*{Learning} & \multirow{2}*{DIST} & \multirow{2}*{Scenario} & \multirow{2}*{Scale} & \multicolumn{4}{c|}{Problems}\\
\cline{12-15}
~&~&~&~&~&~&~&~&~&~&~&Patrol&Coverage&Formation&Navi.\\
\hline
ABBA&\cite{jung1999architecture}&$\surd$&$\surd$&$\surd$&~&~&$\surd$&$\surd$&dynamic&$\leq$ 10&~&~&~&$\surd$\\
\hline
CHARON&\cite{fierro2002framework}&~&$\surd$&~&~&~&~&$\surd$&static&$\leq$ 10&~&~&$\surd$&\\
\hline
Token Passing&\cite{farinelli2006assignment}&~&$\surd$&$\surd$&~&~&~&$\surd$&dynamic&$\leq$ 10&~&~&~&$\surd$\\
\hline
Teamcore&\cite{tambe2000adaptive}&$\surd$&$\surd$&$\surd$&~&$\surd$&~&$\surd$&dynamic&$\leq$ 10&~&~&~&$\surd$\\
\hline
ALLIANCE&\cite{parker1998alliance}&$\surd$&$\surd$&$\surd$&$\surd$&~&~&$\surd$&dynamic&$\leq$ 20&~&~&~&$\surd$\\
\hline
ASyMTRe&\cite{parker2006building}&$\surd$&$\surd$&$\surd$&$\surd$&~&~&$\surd$&dynamic&$\leq$ 20&~&~&~&$\surd$\\
\hline
BITE&\cite{kaminka2005flexible}&$\surd$&$\surd$&$\surd$&~&$\surd$&~&$\surd$&dynamic&$\leq$ 10&~&~&~&$\surd$\\
\hline
DIST Layered&\cite{goldberg2002distributed}&~&$\surd$&$\surd$&$\surd$&~&~&$\surd$&dynamic&$\leq$ 10&~&~&~&$\surd$\\
\hline
STEAM&\cite{tambe1997towards}&~&$\surd$&$\surd$&~&$\surd$&~&$\surd$&dynamic&$\leq$ 100&~&~&~&$\surd$\\
\hline
Market-Based &\cite{dias2000free}&~&$\surd$&$\surd$&$\surd$&~&~&~&dynamic&$\leq$ 20&~&~&~&$\surd$\\
\hline
Hierarchy-Based &\cite{ma2017overview}&~&$\surd$&$\surd$&~&$\surd$&~&~&static&$\leq$ 20&~&$\surd$&$\surd$&\\
\hline
SASS& Ours &$\surd$&$\surd$&$\surd$&$\surd$&~&~&$\surd$&dynamic&$>$100&$\surd$&$\surd$&$\surd$&$\surd$\\
\hline
    \addlinespace[1ex]
\multicolumn{15}{c}{ HET: Heterogeneous,  COOP: cooperation, COM: Communication, NEGO: Negotiation, DEC: Decision Making, DIST: Distributed Agents, Navi: Navigation.}
\end{tabular}}
\end{table*}

\section{Approach Overview}

We design a simple scenario to implement SASS and distributed algorithms. In our scenarios, a group of swarm robots will cooperate to complete some tasks. Since the tasks are dynamically assigned, the robots need to change their plans and adapt to the new scenario to guarantee the group utility. 

In our framework, we decompose the complex tasks into a series of  sub-tasks and recursively achieve those sub-tasks until the entire task is completed. Accordingly,  we can divide the task allocation and execution into three steps: selection, formation, and routing. 
This process can be illustrated as a Behavior Tree \cite{colledanchise2018behavior} that integrates the sense-think-act cycle \cite{mrs2019}. The robots are assumed to have low-level motion control and sensor-based perception system for sensing and navigation.



First, the robots are partitioned into one or more groups to perform multiple tasks such as surveillance and patrolling. Then, they will compute the placement within a formation shape at each task (for simplicity, we assume a circular shape to circle the area of the task location, but other formation shapes can also be considered). Finally, the robots choose a suitable path to get to the goal point in that formation. When the new tasks are assigned, these robots need to split up to form new groups or merge into existing groups. 

Each robot will first verify that there is a task assigned. If assigned, it will compute an appropriate plan according to its current state and needs. It will then communicate with other robots and perform the negotiation and agreement process until there are no conflicts. Finally, the robots will execute their plans. This process is continuously repeated as a loop in a behavior tree, which processes the flow from left to right. 

The proposed framework is formalized using the tuple $M$=$(S$, $A$, $\delta$, $F)$. Here, $S$=$(Pe$, $Pl_1\ldots_n$, $Ne_1\ldots_n$, $A\&E_1\ldots_n)$, where $Pe$ represents perception, $Pl$ represents plan, $Ne$ represents negotiation, and $A\&E$ represents agreement.  The subscript represent the number of iterations for each process, which is finite. $A=(A_1, A_2, \ldots, A_n)$ is a set of individual robot's actions (behaviors). The transition function that maps states (conditions) to actions (behaviors) is $\delta$, defined as $S * A \rightarrow S$. $F$ represent the accept state. 

After the perception phase, a robot could have $n$ plans $Pl = \{Pl_1, Pl_2,\ldots, Pl_n\}$, where each plan depend on each other requiring sequential execution. Therefore, each plan has negotiation $    Ne = \{Ne_1, Ne_2,\ldots, Ne_n\}$ and agreement phases $A\&E = \{A\&E_1, A\&E_2,\ldots, A\&E_n\}$ separately.


To model an individual robot's motivation and needs in the negotiation process, we introduce the priority queue technique inspired by Maslow's hierarchy of human psychological needs \cite{maslow1943theory}. We define a robot's need hierarchy at several levels, as shown in Fig~\ref{fig: needs}. The lowest level represents the robot's safety needs. In all scenarios, a robot should first consider the safety issues (including human-safe operation) like avoiding conflicts or adversarial attacks. When the situation satisfies the robot's safety needs, the robot will consider its basic but vital needs, such as energy and time availability. Then, it will review its capabilities against the task requirements are subsuming the task priority considerations. In the fourth level, the robot finds rewards and utility costs (such as distance and communication) for cooperation within a group. 
The fifth level is reserved for self-upgrade that could potentially allow multi-robot learning strategies. For instance, after finishing the tasks, robots can upgrade their capabilities based on their experiences through the learning and evolution process. It is worth noting that robots' basic and safety needs are flipped compared to the human needs in Maslow's hierarchy.

The needs at the low level are the precondition of entering higher levels. If a robot cannot satisfy its low-level needs, it will change its behaviors accordingly. We design the priority queue to abstract this model and implement it in the negotiation process of our multi-robots planning framework. 
Also, the individual robot's current needs can dynamically change according to different scenarios. For example, in the normal state, the robot's behaviors and plans could maximize the task's requirement and minimize its energy. However, in cases of conflicting plans with other robots (e.g., potential collision with a neighboring robot), it will ensure that the safety needs are guaranteed. Similarly, if the robot runs out of battery, it also needs to find its basic needs first (e.g., recharging battery) before completing it.

\section{Algorithms for Self-Adaptive Swarm System}

SASS considers the following modules: Perception; Communication; Planning; Negotiation; Agreement, and Execution. We will discuss them separately below.

\textbf{\textit{1. Perception:}}
Each robot uses various on-board (local) sensors for localization, mapping, and recognizing objects/obstacles in the environment.



\textbf{\textit{2. Communication:}}
The process of communication between robots includes broadcasting and reception of the robot's messages (state) to/from other robots \cite{tardioli2019pound}. The distributed communication in MRS can be regarded as a connected graph. Each robot keeps on communicating with its adjacent/neighbor's robots and exchanging data until they reach \textit{Information Equilibrium}, which means that every group member has the same information for the entire group (see Alg. \ref{alg:DCM}).

\begin{figure}[tbp]
\centering
\includegraphics[width=0.48\textwidth]{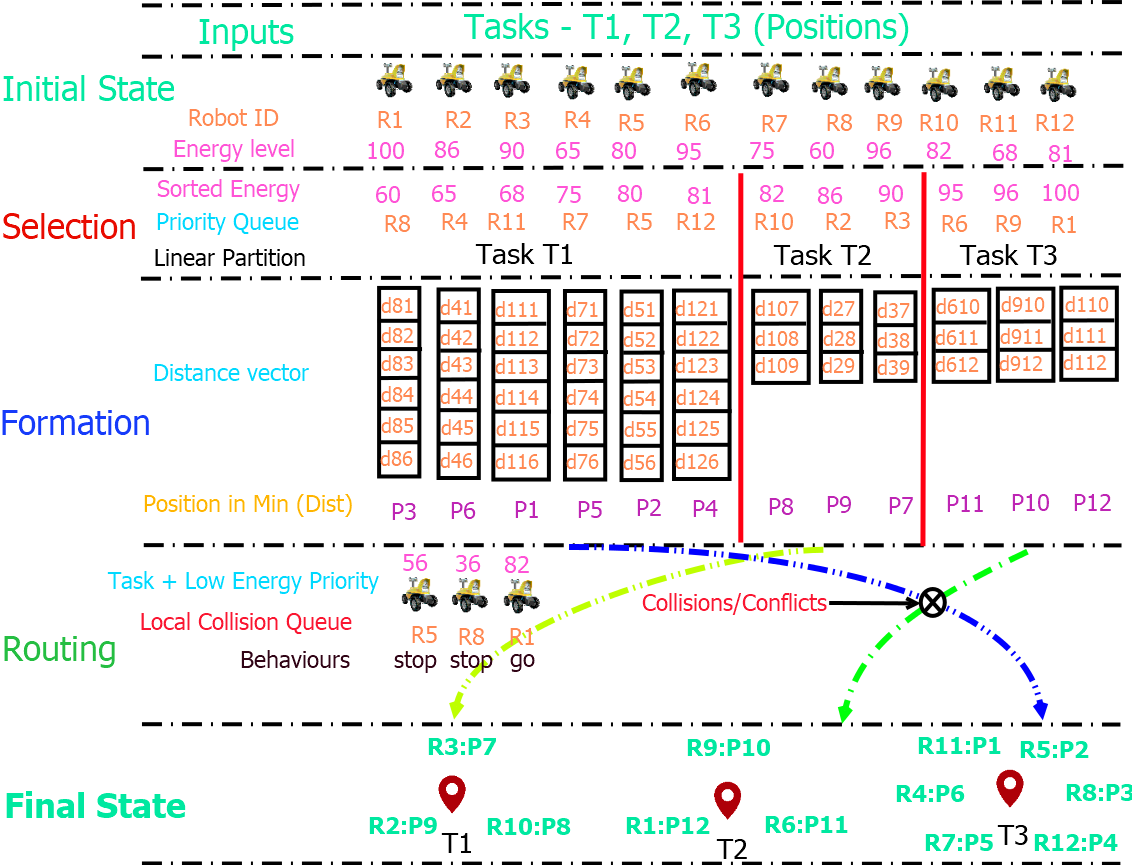}
\caption{Illustration of Task Decomposition through planning at Selection, Formation, and Routing phases with 12 robots and 3 tasks.}
\label{fig:task_decompose}
\end{figure}

\textbf{\textit{3. Planning:}}
In the planning stage, we divide this process into three steps Selection, Formation, and Routing. These \textbf{Atomic Operations} are illustrated in Fig.~\ref{fig:task_decompose} with an example planning problem. In each step, we also introduce a priority queue technique, which can help individual robot negotiate with other robots and get an agreement efficiently.

\textit{3A. Selection Planning:}
In our scenario, we assume that each task has an $ID$ representing its priority (level-4 in Fig.~\ref{fig: needs}), the minimum number of robots required to perform the task, and a task duration (timeout). Also individual robot has an $ID$ and Energy (battery level) (level-2 in Fig.~\ref{fig: needs}). We assume all robots are homogeneous and do not implement the third level of needs (capability). However, for heterogeneous robotic systems, we should consider this priority as well. 

For selection planning, we minimize the sum of the group's energy costs to get a reasonable grouping/partitioning. Then, according to the tasks' priority and requirement, we distribute the diverse tasks to different groups, abstracting it to \textit{The Linear Partition Problem}. 

Through this process, we can ensure that every group can achieve that specific task. Since each robot computing this process in a distributed manner, the local result at the individual robot could potentially have conflicts with the results of other robots depending on the uncertainty in the data it possesses and receives. For instance, one robot might have been assigned to different groups by different robots in the Selection phase. To prevent such a situation, we initiate a negotiation mechanism. We use the priority queue before this process and sort this priority queue with different priority levels, until we get a unique priority queue for all the robots, which can then be used to perform non-conflicting selection planning.

\begin{algorithm}[t]
\SetAlgoLined
\SetKwData{Left}{left}\SetKwData{This}{this}\SetKwData{Up}{up}
\SetKwFunction{Union}{Union}\SetKwFunction{FindCompress}{FindCompress}
\SetKwInOut{Input}{Input}\SetKwInOut{Output}{Output}
\SetNlSty{textrm}{}{:}
\SetKwComment{tcc}{/*}{*/} 

\scriptsize{

\Input{
		Robot $i$ data $d_i$, The number of robots in group $n$ and Adjacent Robots Set $r_a$
       }
\Output{
		The Information Equilibrium data set $D_i$
}
\BlankLine
initialization\;
    memory data set $D_i$ = \{$\varnothing$\} \\
    $D_i$.add($d_i$) \\
    \While{length($D_i$) $\neq$ n}
    {    \For{each j $\in$ $r_a$}
        {
            $D_i$ = $D_i$ $\cup$ $D_j$
        }
    }
} 

\caption{\footnotesize{Distributed Communication Mechanism (DCM)}}
\label{alg:DCM}
\end{algorithm}

\begin{algorithm}[t]
\SetAlgoLined
\SetKwData{Left}{left}\SetKwData{This}{this}\SetKwData{Up}{up}
\SetKwFunction{Union}{Union}\SetKwFunction{FindCompress}{FindCompress}
\SetKwInOut{Input}{Input}\SetKwInOut{Output}{Output}
\SetNlSty{textrm}{}{:}
\SetKwComment{tcc}{/*}{*/} 

\scriptsize{
\Input{		Unsorted priority queue $q_i$, potential collision queue $q_c$,       }
\Output{
		Sorted priority queue $q_{si}$
}
    \If{State == Selection/Formation/Routing}
    {
	    $q_n$ = hierarchical needs order queue;
	}
	\If{State == Routing}
    {
        $tmp1$ = DCM($q_c$)\;
        \For{each $item$ in Union-Find($tmp$)}
        {   
            \If{i $\in$ $item$}
            {$q_i$ = $item$}
        }
    }
    \If{$q_n~!= NULL$}
    {
        $q_{si}$ = Sort $q_i$ with $q_n.first$\;
        $tmp2$ = DCM($q_{si}$)\;
    	\While{$Agreement(tmp2) == "conflict"$}
    	{
    	    $q_{si}$ = Sort $q_i$ with $q_n.next$\;
    	    $tmp2$ = DCM($q_{si}$)\;
    	}
    }
} 

\caption{\small{Selection/Formation/Routing Negotiation}}
\label{alg:SPN}
\end{algorithm}

\begin{algorithm}[h]
\SetAlgoLined
\SetKwData{Left}{left}\SetKwData{This}{this}\SetKwData{Up}{up}
\SetKwFunction{Union}{Union}\SetKwFunction{FindCompress}{FindCompress}
\SetKwInOut{Input}{Input}\SetKwInOut{Output}{Output}
\SetNlSty{textrm}{}{:}
\SetKwComment{tcc}{/*}{*/} 

\scriptsize{

\Input{
		Sorted priority queue list $Q$
       }
\Output{
		Execute the plan or negotiation again
}
\BlankLine
initialization\;
    count = 1\;
    \For{each $item$ in $Q$}
    {
        \If{$Q.first$ != $item$}
        {count++}
    }
    \eIf{count == 1}
    {
        return ``end'' and execute the corresponding ``Atomic Operation''\;
    }
    {
        return ``conflict''.
    }
} 

\caption{\small{Selection/Formation/Routing Agreement}}
\label{alg:SA}
\end{algorithm}

\textit{3B. Formation Planning:}
Each robot will make a formation plan according to the selected plan. Here, each robot knows which group (task) it belongs to. To simplify the formation models, we assume the robots need to create a regular polygon surrounding the task assignment location (group's center). The initial point will be located at the North position and follow the clockwise order to arrange the other point assignments within the formation by minimizing the system utility as mentioned before (we use distance as utility cost and energy level for prioritizing the robot's needs).


\textit{3C. Route Planning:}
Each robot computes the routes (path plan) using the local environment map of the sensor data and selects the shortest path getting to the goal point resulting from the Formation plan. Suppose each robot has two kinds of actions/behaviors that can be selected: one is moving at a constant speed($v$), the other is stop. In case some robots have a conflict in the process of making the routing plan, they will create a priority queue with all conflicting robots $ID_{i \ldots j}$ and share with the neighbors through local communication. Then, according to the Task and Energy priority of the robots, each robot in this queue will negotiate to decide on the corresponding actions on the robots that have conflicts. Until an agreement is reached, the priority queue is updated based on the needs hierarchy and solve the conflict. 

\textbf{\textit{4. Negotiation:}}
Robots will compare the plans received from other group members with their own. 
For the Selection and Formation plan, the negotiation will be performed until the robots are assigned to only one group (in case of selection) or one position in the formation. 
Route planning involved creating a unique priority queue that avoids conflicts. Subsequently, each robot will reach an agreement on the priority queue and the corresponding plans. 
For example, in the selection and formation section, we use Low Energy ($ Low_E $), which means the robots with lower battery levels get higher priority and similarly the High Energy $ High_E $ law. We combine them with using Task-based priority queue (each task will have a specific priority in assigning the tasks), resulting in $T+High_E$ and $T+Low_E$. We also consider whether or not to consider the priority needs of conflicts for route planning (conflict case and No conflict case).
We present the algorithmic representation of the negotiation process in Alg.~\ref{alg:SPN}.

For instance, in the Formation plan negotiation, we use distances between its local position and the task's polygon points in the boundary of the formation shape. Each robot communicates this distance vector with other robots in the same group assigned to this specific task. Each robot will compute a matrix (Level 4 Team needs - Utility cost)  representing each robot's distances to all the polygon points in the specific task. Then it will use the unique queue order to select the corresponding distance until all the group member gets the specific task goal point as long as the priority queues of Task, Energy, and Safety are satisfied. For example, the low energy robot will have a high priority choosing the point closest to it, thereby reducing energy consumption.


\textbf{\textit{5. Agreement and Execution:}}
If all the robots' plans do not have conflict after the negotiation phase, they will have a final agreement per process (Selection/Formation/Routing). The algorithm for implementing the agreement process is presented in Alg.~\ref{alg:SA}.
After the agreement, each process is executed as and when necessary.

\begin{figure*}[t]
\centering
\subfigure[Comparison of conflict frequency.]
{
	\centering
	\includegraphics[width=0.315\textwidth]{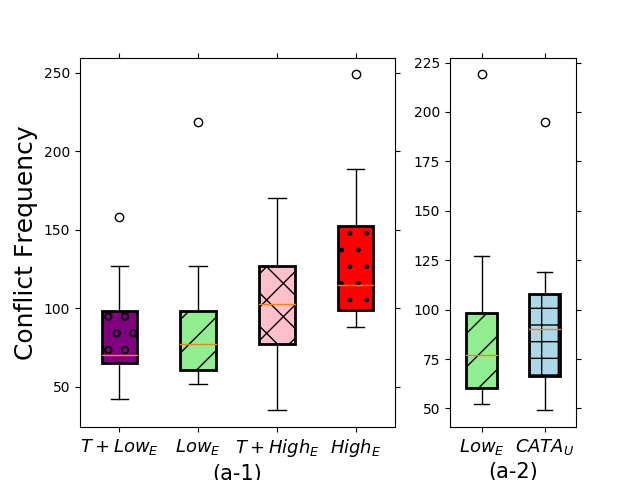}
	\label{fig: collision20-s}
}
\subfigure[Comparison of total energy costs.]
{
	\centering
	\includegraphics[width=0.315\textwidth]{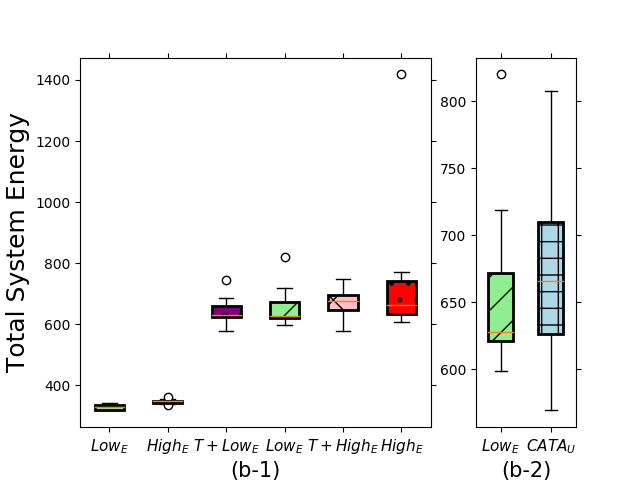}
	\label{fig: energy20-s}
}
\subfigure[Comparison of total travel distance.]
{
\centering
\includegraphics[width=0.315\textwidth]{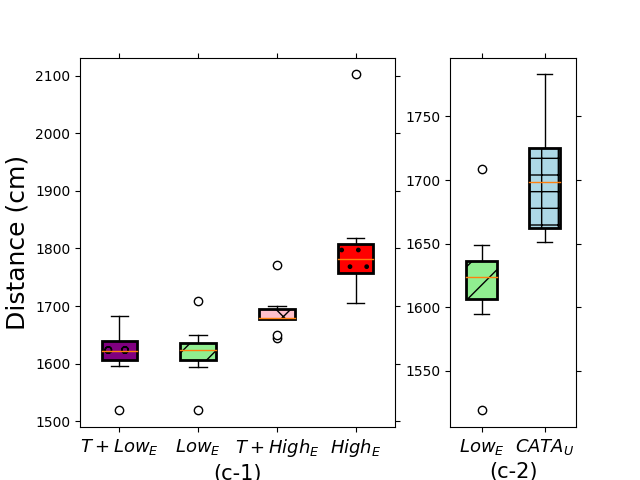}
\label{fig: distance20-s}
}
\caption{Experiments on static task assignments with 20 robots and 3 tasks.}
\vspace{-2mm}
\end{figure*}

\section{Evaluation through Simulation Studies}

To simulate our framework, we chose to use the ``Common Open Research Emulator (CORE)'' network simulator \cite{ahrenholz2010comparison} since we are interested in implementing our algorithm in a network-based tool as CORE allows dynamic changes in the node/agent mobility and communication. 
We consider 20 robots in our simulations due to limitations in the CORE framework for an illustration of a single task assignment with 20 fully connected robots). In the evaluations, we consider only the lower three levels (Safety + Basic + Capability) in the robot needs hierarchy (see Fig.~\ref{fig: needs}) for aiding rigorous analysis and validation of our cooperation framework. Composition of higher level needs will be investigated in our future work. 

We suppose each robot has different battery levels in the initial state, and every moving step will cost $0.1\%$ energy. Also, every communication round and non-moving status will cost $0.01\%$ and $0.04\%$ energy, respectively. To simplify the visualization of the utility of the framework, we do not consider any obstacles. We design two scenarios – one to simulate static task assignments (all tasks are added at the initial stage) and another to simulate a dynamic task assignment (a task can be added anytime during the process).

We consider four combinations of priority: $High_E$ (High Energy), $Low_E^{Num}$ (Low Energy + Task Priority Order), $T+High_E$ (Task Priority + High Energy), and $T+Low_E$ (Task Priority + Low Energy). 
For example, if we adopt a priority queue with task + low energy combination, the scenario would be to first address the emergency task and maintain robots in the field as long as possible. We intend to compare the utility and behaviors of the individual robot and the system with different priority combinations.

To compare our approach with a state of the art method, we implemented the algorithm called Collision-Aware Task Allocation (CATA) in \cite{wu2019collision} in which the authors proposed a new method for addressing collision-aware task assignment problem using collision cone and auction-based bidding algorithms to negotiate the conflicts. Since our framework is distributed and does not have a central agent to manage the bidding process, we implemented the algorithm in \cite{wu2019collision}, which provides the rewards for each robot to each task location. The rewards are converted to a Utility matrix and fed to the negotiation mechanism in our framework. Therefore, we term this method as $CATA_U$ (Collision-aware Task ASsignment + Utility Matrix). Here, a robot first calculates the task's utility based on \cite{wu2019collision}, and it chooses the maximum utility task based on the low energy priority law. Therefore, we compare this method only with our $Low_E$ priority law.
The experiment demonstrations are available online\footnote{Experiment demonstration video is available at the website: \url{https://hero.uga.edu/research/sass}}.


\begin{figure*}[t]
\centering
\subfigure[Comparison of conflict frequency.]
{
	\centering
	\includegraphics[width=0.315\textwidth]{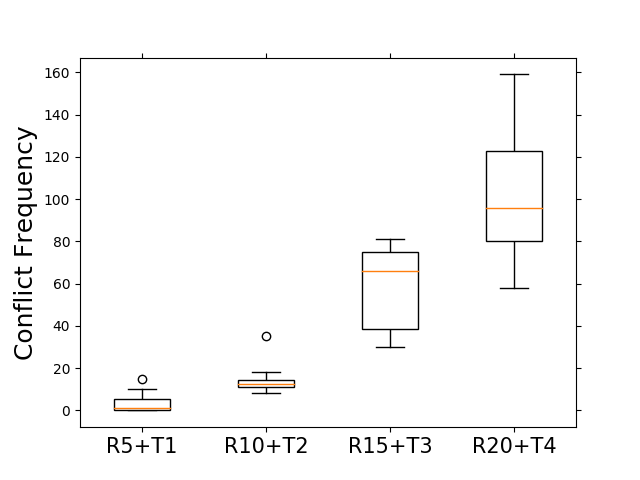}
	\label{fig: collision_frequency}
}
\subfigure[Comparison of Comm. \& moving energy costs.]
{
	\centering
	\includegraphics[width=0.315\textwidth]{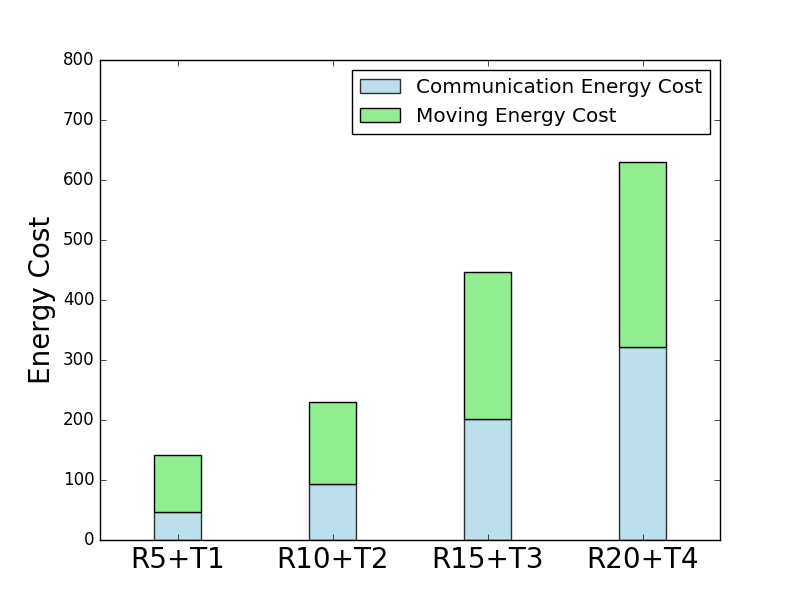}
	\label{fig: negotiation_vs_moving_energy}
}
\subfigure[Comparison of per task Comm. energy costs.]
{
\centering
\includegraphics[width=0.315\textwidth]{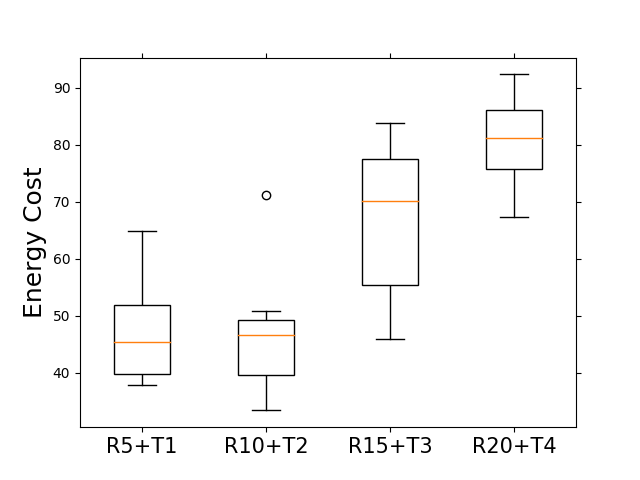}
\label{fig: per_task_negotiation_energy_cost}
}
\caption{Experiments on Scalability and Complexity in Static Task Assignments. Here, R15 means 15 Robots and T4 means 4 Tasks for example.}
\vspace{-2mm}
\end{figure*}

\begin{figure*}[t]
\centering
\subfigure[Comparison of conflict frequency.]
{
	\centering
	\includegraphics[width=0.315\textwidth]{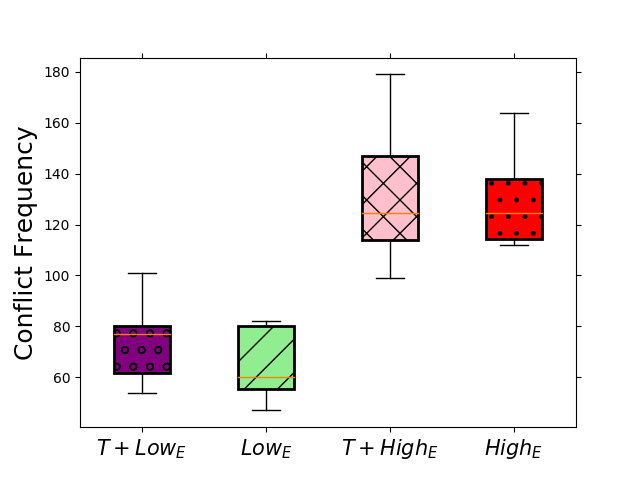}
	\label{fig: collision_frequency_p}
}
\subfigure[Comparison of Comm. energy costs.]
{
	\centering
	\includegraphics[width=0.315\textwidth]{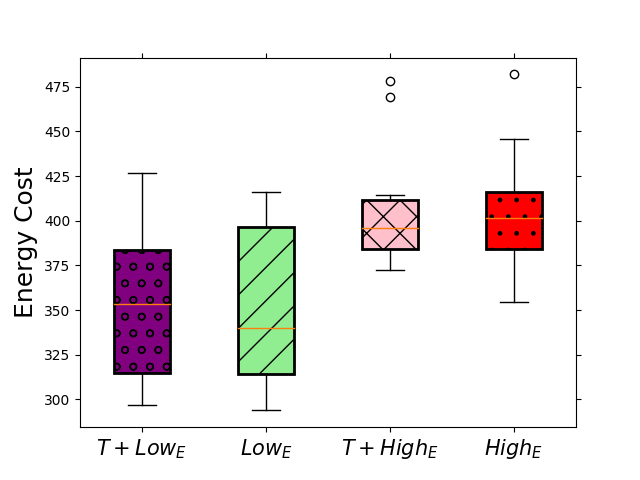}
	\label{fig: communication_energy_cost}
}
\subfigure[Comparison of Low$_E$ and CATA$_U$.]
{
\centering
\includegraphics[width=0.315\textwidth]{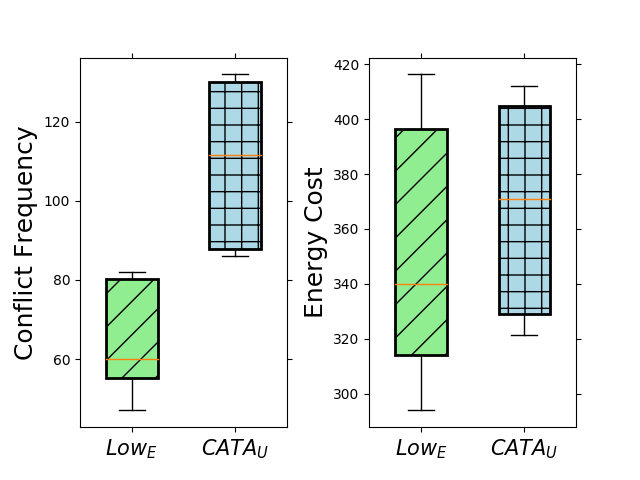}
\label{fig: system_energy_cost}
}
\caption{Experiments on the Impact of Different Priority Laws in Static Task Assignments.}
\label{fig:priority}
\vspace{-2mm}
\end{figure*}

\subsection{Static Multi-Task Assignments}
We conduct ten simulation trials for each priority case in a static task assignment scenario. In every priority case, we use the same ten different initial battery levels sampled from a Gaussian distribution with a mean of $90\%$ and a standard deviation of $10\%$. 
Fig. \ref{fig: collision20-s} and \ref{fig: energy20-s} shows the distance matrix results of four priority laws of conflict frequency with and without conflict negotiation costs comparing the CATA approach. In this experiment, the $T+Low_E$ priority combination had the best performance compared with other cases. At the same time, Fig. \ref{fig: energy20-s} shows that every group almost cost one-third of the entire energy in the negotiation and agreement part of solving the conflicts. This means that more conflicts will lead to more negotiation rounds and corresponding energy consumption in communication. 
In Fig. \ref{fig: distance20-s}, the effect of each priority law in the total system distance is also demonstrated for our finding that different needs and priorities at the individual agent level will lead to various global performances.

In the priority law of $ Low_E $, our method had fewer conflicts than the $ CATA_U $ method. We believe this is because prioritization of basic needs in our approach aims to avoid conflicts at the Formation planning stage. On the other hand, $CATA_U$ considers the conflicts only at the Route planning stage, which would leave more room for conflicts if the task polygon points are not efficiently assigned in the formation stage itself.

\subsection{Scalability and Complexity}

\begin{figure*}[htbp]
\centering
\subfigure[Comparison of conflict frequency]
{
	\centering
	\includegraphics[width=0.313\textwidth]{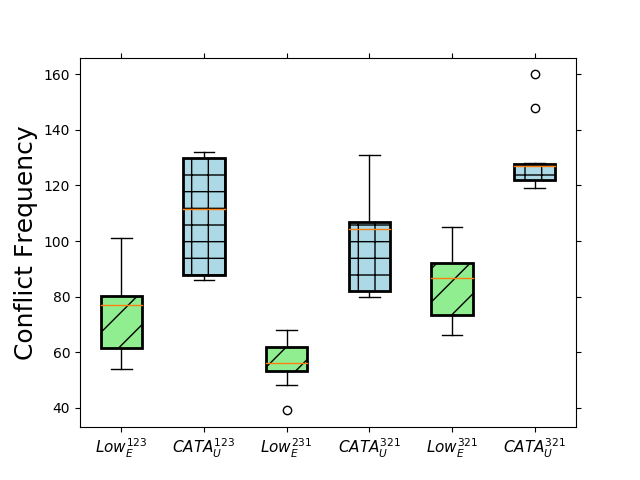}
	\label{fig: dis_vs_qap_f}
}
\subfigure[Comparison of total energy cost.]
{
	\centering
	\includegraphics[width=0.313\textwidth]{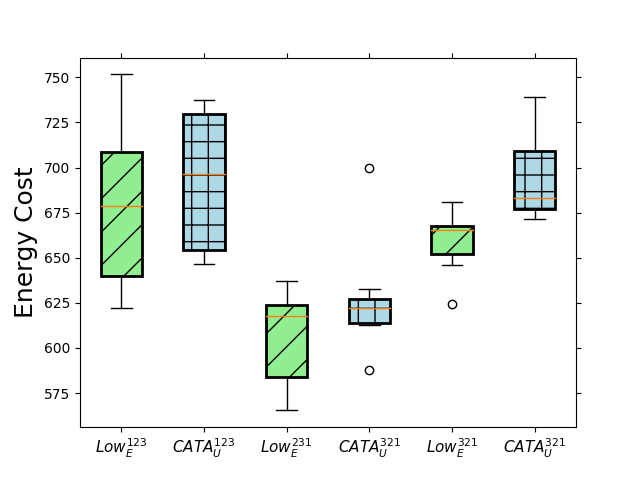}
	\label{fig: dis_vs_qap_e}
}
\subfigure[Comparison of total battery level.]
{
	\centering
	\includegraphics[width=0.32\textwidth]{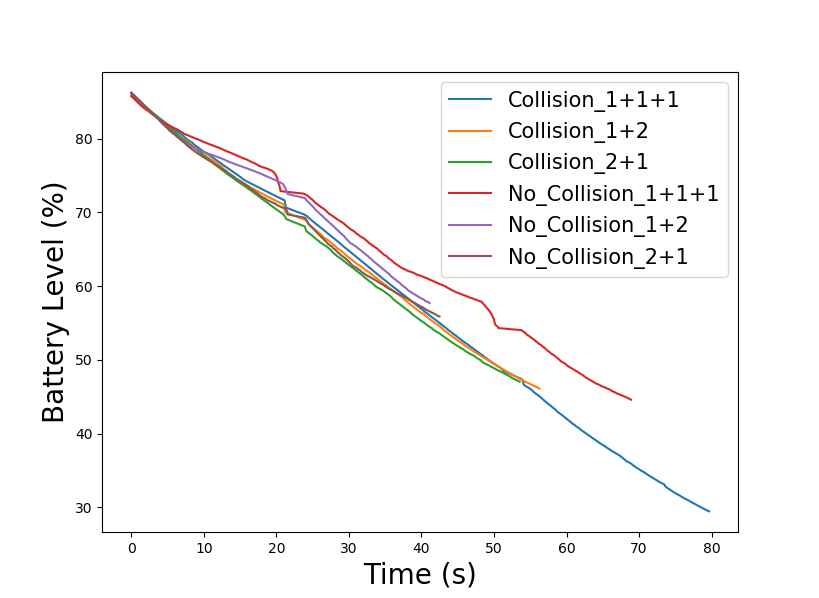}
	\label{fig: energy20-d2}
}
\caption{Experiments on the impact of different priority laws and dynamic task assignments with 20 robots and 3 tasks. (Left and Center) Comparison of conflict frequency and energy cost based on the different task's priority in statics task assignments. E.g., In $Low_E^{123}$, the priority of T1 $>$ T2 $>$ T3. (Right) Comparison of the total battery level of the system measured in points combining all robot energy percentage levels in dynamic task assignments.}
\end{figure*}

To verify SASS's scalability and the complexity of the MRS cooperation, we design four scales of robots' team implementing in different numbers of tasks: $R5+T1$, $R10+T2$, $R15+T3$, $R20+T4$. For example, $R10$ means ten robots in the system, and $T3$ means three different tasks are assigned. 

Through Fig. \ref{fig: collision_frequency}, we can notice that as the number of robots and the task complexity increase, the entire system's conflicts (conflict frequency) rise rapidly. The proportion of communication energy cost compared with the moving energy cost also increase with the increase in scale (see Fig. \ref{fig: negotiation_vs_moving_energy}). This points to the fact that the whole system spends more energy and time negotiating and cooperatively converging to a specific agreement. So considering the average communication energy cost per task (see Fig. \ref{fig: per_task_negotiation_energy_cost}), if the task complexity is higher in some specific scenarios or the environment is more unstructured and unpredictable, an individual agent will spend more energy and time in communication to fulfill the tasks.

From another perspective, the fully distributed communication graph is inefficient in a large scale system of coordinating robots, especially in the swarm robots. Therefore, designing a proper communication architecture to adapt to a particular scale of agents' group and complexity scenarios is also an important and challenging problem that should be investigated further, which is an avenue for future work.

\subsection{Impact of Different Priority Laws at Different Levels}
If an individual agent has different needs or motivations, it might present various conduct, leading to the entire system displaying different performance. To verify this hypothesis, we use 20 robots having the same initial battery levels based on four different priority laws comparing with the distance matrix and CATA as we discussed above. We conduct ten trials with different initial battery levels sampled from a Gaussian distribution with a mean 90\% and standard deviation 30\% to represent heterogeneity in the energy capacities of the robots.

Fig.~\ref{fig: collision_frequency_p} shows that implementing different priority laws causes the whole system to exhibit different energy use. As we can see, the system has better performance in $ T+Low_E $ and $ Low_E $ priority laws than $ T+High_E $, $ High_E $ priority laws. This is because the former two lead to fewer conflicts than the latter two. 
We can confirm these observations through Fig. \ref{fig: communication_energy_cost} and \ref{fig: system_energy_cost}.

We also shuffle the task priorities (e.g., the priority of T1 $>$ T2 $>$ T3 in the superscript 123) to simulate the variation in the third level of the needs hierarchy (Capability/Requirements) and the results are shown in Fig. \ref{fig: dis_vs_qap_f} and Fig. \ref{fig: dis_vs_qap_e}. We can see that our approach's distance matrix always has better performance than the CATA under the same conditions in terms of conflict resolution and system energy.

In practical applications, an intelligent agent might dynamically change its needs or motivation according to the situations, especially in the adversarial or unpredictable environment. This may lead to chaos in the system resulting in higher energy costs and loss of system utility. In the MRS design, letting the individual agents select suitable laws for coordination while guaranteeing the optimal system utility is also an exciting and challenging avenue for future work.


\subsection{Dynamic Multi-Task Assignments}
\label{sec:dynamic}

In the dynamic multi-tasks scenario, we design three kinds of dynamic tasks. One is three tasks added sequentially after every task is completed ($1+1+1$). The rest of the two cases are two tasks appearing at the start ($2+1$) and the end ($1+2$) in the entire process. Also, we combine this with a conflict or no conflict cases.


\begin{table}[tbp]
\begin{center}
\caption{Energy Level Comparison in Dynamic Task Assignments}
\scriptsize
\begin{tabular}{cccccc} 
\hline 
Tasks Style&Priority&conflict&Max&Min&Mean\\
\hline
1+1+1&High$_E$&$\surd$&69.42&55.52&64.18\\
\hline
1+1+1&Low$_E$&$\surd$&63.09&45.70&56.00\\
\hline
1+1+1&T+High$_E$&$\surd$&70.04&56.75&63.24\\
\hline
1+1+1&T+Low$_E$&$\surd$&63.25&49.18&56.62\\
\hline
1+2&T+Low$_E$&$\surd$&45.97&31.78&39.60\\
\hline
2+1&T+Low$_E$&$\surd$&44.69&31.66&39.07\\
\hline
1+1+1&T+Low$_E$&-&49.88&29.48&41.20\\
\hline
1+2&T+Low$_E$&-&32.70&23.14&28.50\\
\hline
2+1&T+Low$_E$&-&38.53&24.72&30.36\\
\hline
\end{tabular}
\label{table: energy20-d}
\end{center}
\vspace{-2mm}
\end{table}

The first experiment we consider using four different priority laws and 20 robots implementing the $1+1+1$ scenario (see Table \ref{table: energy20-d}). Here, the initial energy level and position for each priority laws were set the same. We can observe that the $T+Low_E$ and $Low_E$  combinations achieved the best system utility as in the static assignment cases. 

In the second experiment, we evaluate the impact of the conflict negotiation process by comparing the energy costs of three different tasks with and without considering conflicts. In Fig. \ref{fig: energy20-d2}, we notice that the negotiation cost occupies a large part in the entire system costs (hence, the higher battery level consumed when collisions are considered). We can also observe that the difference between each combination's negotiation energy cost reflects the environment's unstructured level, which means the difference will increase in the more chaotic and dynamic considerations.

\section{Conclusion}

Our work introduces a novel SASS framework for cooperation heterogeneous multi-robot systems for dynamic task assignments and automated planning. It combines robot perception, communication, planning, and execution in MRS, which considers individual robot's needs and action plans and emphasizes the complex relationships created through communication between the robots. Specifically, we proposed \textit{Robot's Needs Hierarchy} to model the robot's motivation and offer a priority queue in a distributed \textit{Negotiation-Agreement Mechanism} avoiding plan conflicts effectively. Then, we provide several \textit{Atomic Operations} to decompose the complex tasks into a series of simple sub-tasks. 
The proposed solution is evaluated through extensive simulations under different static and dynamic task scenarios. The experimental analysis showed that the needs-based cooperation mechanism outperformed state-of-the-art methods in maximizing global team utility and reducing conflicts in planning and negotiation. 



SASS leaves room for many future improvements. 
For instance, we plan to optimize the communication architecture and add decision learning levels into individual robot's hierarchical needs completing our robot's needs model. This improvement will cause the individual robots to upgrade by itself based on the learned experiences and lead to the self-evolution of the whole system.

\bibliographystyle{IEEEtran}
\bibliography{references}



\end{document}